\pdfoutput=1

\documentclass[11pt]{article}

\usepackage{acl}

\usepackage{times}
\usepackage{latexsym}

\usepackage[T1]{fontenc}

\usepackage[utf8]{inputenc}

\usepackage{microtype}
\usepackage{amsmath}
\usepackage{amssymb}
\usepackage{graphicx}
\usepackage{multirow} 
\usepackage{multicol}
\usepackage{makecell}
\usepackage{tabularx}
\usepackage{color}
\usepackage{xcolor}
\usepackage{listings}
\usepackage{subcaption}
\usepackage{url}
\usepackage{latexsym}
\usepackage{booktabs}
\usepackage{adjustbox}
\usepackage{url}
\usepackage{xspace}
\usepackage{pgfplots}
\usepackage{tikz}
\usepackage{xargs}
\usepackage{diagbox}
\usepackage{adjustbox}
\usetikzlibrary{positioning}
\usepackage{pifont}
\usepackage{float}
\restylefloat{table}
\newcommand{\cmark}{\ding{51}}%
\newcommand{\xmark}{\ding{55}}%

\newcommand{\ignore}[1]{}

\newcommand*\samethanks[1][\value{footnote}]{\footnotemark[#1]}

\title{Domain-matched Pre-training Tasks for Dense Retrieval}
\author{
Barlas Oğuz\thanks{\hspace{.06in}Equal contribution}, Kushal Lakhotia\samethanks, Anchit Gupta\samethanks,  \\
\textbf{Patrick Lewis, Vladimir Karpukhin, Aleksandra Piktus, Xilun Chen,} \\
\textbf{Sebastian Riedel, Wen-tau Yih, Sonal Gupta, Yashar Mehdad} \\
\\
Facebook AI Research \\
\\
{\tt \{barlaso,kushall,anchit,plewis,vladk,piktus,} \\
{\tt xilun,sriedel,scottyih,sonalgupta,mehdad\}@fb.com} \\
}

\date{}

\begin{document}
\maketitle
\begin{abstract}
Pre-training on larger datasets with ever increasing model size is
now a proven recipe for increased performance across almost all NLP tasks.
A notable exception is information retrieval, where additional pre-training
has so far failed to produce convincing results.  We show that, with the
right pre-training setup, this barrier can be overcome.  We demonstrate this
by pre-training large bi-encoder models on 1) a recently released set of 65 million
synthetically generated questions, and 2) 200 million post-comment pairs from a preexisting dataset of Reddit conversations made available by pushshift.io.
We evaluate on a set of information retrieval and dialogue retrieval benchmarks, 
showing substantial improvements over supervised baselines.

\end{abstract}

\section{Introduction}
As a pre-training task, language modeling and its variants (causal~\citep{radford2018improving}, bi-directional~\citep{peters2018deep, baevski2019cloze}, masked~\citep{devlin2018bert}, seq2seq~\citep{bart, t5}) have proven to be extremely versatile and shown to transfer well to most, if not all NLP tasks.  Nevertheless, in-domain fine tuning remains important, as there is still a gap between the pre-training task and the downstream tasks.  Numerous approaches have been proposed to fill this gap, with an additional (intermediate) pre-training stage, mostly based on multi-task learning~\citep{t5, aghajanyan2021muppet}.  It's been generally accepted that the more similar the end task is to the pre-training task, the larger the gains (e.g., NLI tasks transfer better to other NLI tasks~\citep{phang2018sentence}, QA tasks to QA tasks~\citep{khashabi2020unifiedqa}, \textit{inter alia}).

From this perspective, information retrieval (IR), which is the task of identifying the most relevant document to a given query from a large corpus of candidates, has a unique position.  At the surface, IR looks similar to other NLP tasks in standard benchmarks, such as NLI or paraphrase detection.  However, the need to accommodate large corpora imposes computational constraints, which lead to important practical differences.  Most importantly, indexing needs to happen offline, therefore the candidate representations need to be calculated independently of the query representation.  As a result, neural retrieval systems typically use a \textit{bi-encoder} model (Figure~\ref{fig:biencoder}), trained to minimize the similarity between the document representation and the query representation.  This shallow interaction between document and query encoders makes neural IR models architecturally unique, compared to the block cross-attention transformers which are the universal choice for almost every other NLP task.

Researchers have therefore recognized the need to construct intermediate pre-training tasks, that are better matched to retrieval.  \citet{orqa} proposed the \textit{inverse cloze task (ICT)}, which treats sentences as pseudo-queries, and matches them to the paragraph they originate from. \citet{chang2020pre} combined this with \textit{body first selection (BFS)} (selecting the first paragraph given a sentence from the same document), and \textit{wiki link prediction}.  \citet{REALM} pre-trained a retrieval model jointly in an end-to-end system to minimize a language modelling objective.  

In each of these cases, pre-training approaches were shown to improve over their respective baselines.  However, subsequent work showed that a careful fine-tuning over a vanilla BERT model can outperform all of these approaches~\cite{dpr}.  The findings for model scaling are also similar to those of data scaling.  Published results show only modest improvements from larger models for retrieval, and retrieval models which top the most competitive document ranking leaderboards are still based on the relatively small BERT-base architecture.\footnote{https://microsoft.github.io/msmarco/}  This is in sharp contrast to other NLP benchmarks, where data and model scaling has been extremely successful. 

We hypothesise that previously proposed pre-training tasks might be still too distant from the target task, which limits useful transfer.  We therefore investigate pre-training tasks for retrieval which are as closely matched to the the target task and domain as possible.  To this end, we propose using two corpora for retrieval pre-training:
\begin{itemize}
  \item A corpus of 65 million synthetically generated question-answer pairs from Wikipedia~\citep[PAQ,][]{lewis2021paq}, which we target for open domain question answering and other passage retrieval tasks.
  \item A corpus of 220 million post-comment pairs from Reddit, which we use for dialogue retrieval tasks.
\end{itemize}

We conduct extensive evaluations on two popular information retrieval tasks, a benchmark composed of 8 knowledge-intensive retrieval tasks, and 3 dialogue retrieval benchmarks.  We find that pre-training leads to strong improvements in all cases, and also demonstrate robust generalization.  We compare different pre-training tasks, investigating the effect of domain and task similarity, and find both to be important.  We also experiment with models of varying sizes, with and without pre-training, showing in some cases that retrieval can indeed benefit from larger models.

\section{Dense retrieval}
In this section we give an overview of dense retrieval models and how they are trained.

\subsection{Bi-encoder architecture}
\begin{figure}
    \centering
    \includegraphics[scale=0.4]{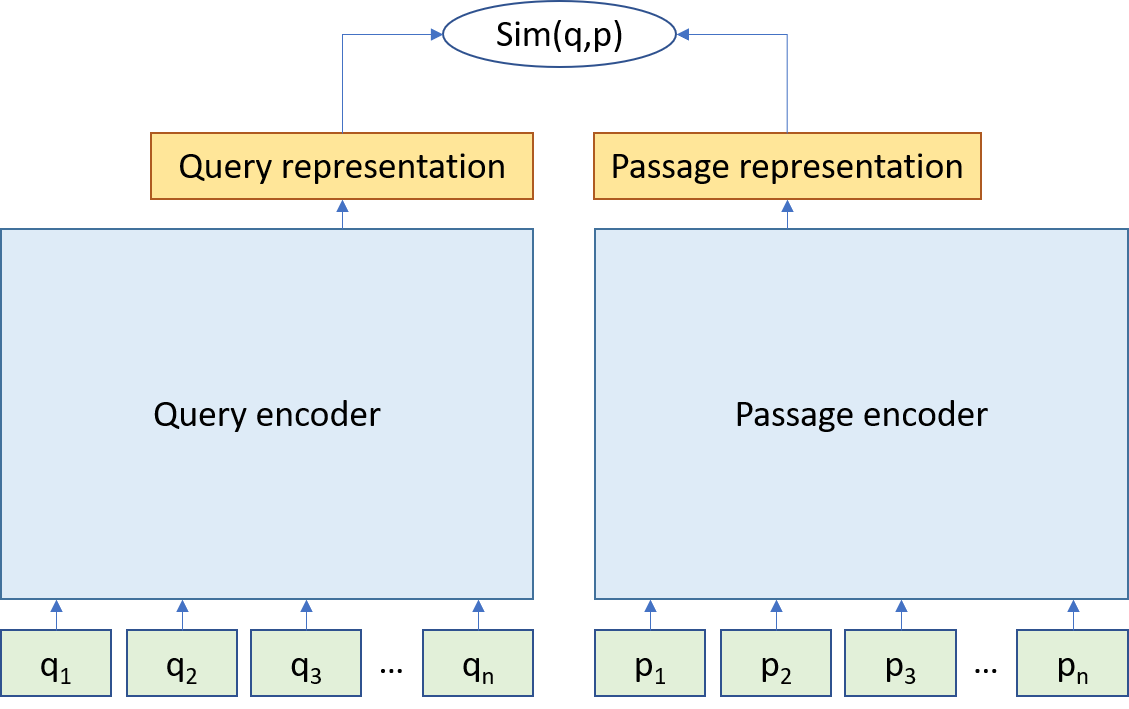}
    \caption{Bi-encoder architecture for retrieval.}
    \label{fig:biencoder}
\end{figure}
A typical dense retrieval system consists of a query encoder $E_Q$, a passage encoder $E_P$ which output a fixed $d$-dimensional representation for each query and passage respectively.  Passages are processed offline, and their representations are indexed using a fast vector similarity search library such as FAISS~\citep{faiss}.  At runtime, an incoming query is encoded, and the top-$k$ closest passages to its representation in vector distance are returned using the index.  Dot-product similarity is most commonly used:
\begin{align}
    \mathrm{sim}(q, p) = E_Q(q)^{\intercal} E_P(p).
    \label{eq:sim}
\end{align}
The resulting \textit{bi-encoder} architecture is pictured in Figure~\ref{fig:biencoder}.  Crucially, this formulation allows passage representations to be calculated independently from the query encoder, making efficient retrieval possible.

\subsection{Training}  
Given a query, a relevant (positive) passage, and a list of non-relevant (negative) passages, the bi-encoder model is trained to minimize the negative log likelihood of picking the positive passage, where the probability assigned to each passage is proportional to $e^{\mathrm{sim}(q, p)}$.  For efficiency reasons, positive passages are recycled as negative passages for queries they are not paired with in the batch, referred to as \textit{in-batch} negatives.  In addition, \textit{hard} negatives have been found to be useful, which can either come from a standard retrieval system such as BM25, or an earlier iteration of the dense model~\citep{xiong2020approximate}.  We do training in two steps (which we refer to as \textit{iterative training}).  In the first step we use a single BM25 negative per query, following best practice from~\citet{dpr}, and in the second step we use hard negatives obtained using the first round model. This procedure approximates the asynchronous model update which was shown to be helpful in~\citet{xiong2020approximate}.

\section{Experimental setup}

\subsection{Pre-training tasks}
In this section, we describe the datasets we used to pre-train our retrieval models.
\subsubsection{PAQ}
For open-domain question answering tasks, we employ the recently-released PAQ dataset~ \citep{lewis2021paq}. 
This dataset consists of 65 million synthetic question-answer pairs, generated from Wikipedia passages.
PAQ is generated by a pipeline of models trained on Natural Questions~\cite[NQ][]{nq} and TriviaQA \cite[TQA][]{joshi-etal-2017-triviaqa}.
%
PAQ's main distinguishing features relative to other QA-pair generation techniques are its large size, and the use of a novel \emph{global} consistency filtering. 
This leads to higher quality, less ambiguous open-domain-style questions than can be achieved standard consistency filtering with machine-comprehension models~\cite{alberti_synthetic_2019}.

PAQ has previously been employed as a semi-structured knowledgebase of facts extracted from wikipedia, and used as data-augmentation for closed-book question answering models~\cite{t5close}.
However, since QA-pairs in PAQ are generated from wikipedia passages, we can repurpose PAQ as a source of training data for a passage retrieval task.
Here, given a PAQ question, the task is to retrieve the wikipedia passage that was used to generate said question from a pool of negatives.
PAQ's size makes this a suitable large-scale pretraining task and represents a close proxy of the actual downstream open-domain QA retrieval task.

\subsubsection{Reddit}
For dialogue tasks, we use 200 million post-comment pairs mined from Reddit.  This dataset was originally extracted and made available by \texttt{pushshift.io} and shown to be useful for dialogue and chit-chat applications previously~\citep{humeau2019poly,roller2020recipes}.

\subsection{Evaluation tasks}
In this section, we describe our evaluation setup.
\subsubsection{Passage retrieval}
We evaluate on a mix of standard information retrieval and open-domain question answering benchmarks, and a suite of knowledge-intensive retrieval tasks:
\paragraph{MSMARCO}~\citep{nguyen2016ms} is a suite of benchmarks created using real user queries to the Bing search engine, with human annotated search results.  We evaluate on the passage retrieval task, which is widely reported on in the IR community.
\paragraph{Natural Questions}~\citep[NQ,][]{nq} is a popular open-domain QA dataset, with questions originating from Google users, and answers annotated from Wikipedia.
\paragraph{KILT}~\citep{petroni2020kilt} is a benchmark consisting of a diverse set of 8 knowledge-intensive tasks, including fact-checking, entity linking, relation extraction, dialogue and question answering.  All tasks are grounded in Wikipedia, and we report on the passage selection metrics.

\subsubsection{Retrieval for dialogue}
We also evaluate on a set of dialogue retrieval benchmarks, to show the generality of our conclusions.
\paragraph{ConvAI2} is based on the PersonaChat dataset~\citep{zhang2018personalizing}, and was presented for the Neurips ConvAI2 competition~\citep{dinan2019second}. The task involves selecting the correct next utterance in a dialogue, out of 20 candidates, given the dialogue history as well as some context about the speakers persona.
\paragraph{Ubuntu v2}~\citep{lowe2015ubuntu, arxiv18disentangle} is a large corpus of 1 million conversations from Ubuntu chat logs, which document users receiving support from other users regarding Ubuntu-related issues.
\paragraph{DSTC7}~\citep{gunasekara2019dstc7} is a challenge set consisting of 100k samples extracted from the Ubuntu dataset described above.

\subsection{Implementation}
\paragraph{Frameworks}
We use the Pytorch Lightning (PL) framework~\citep{falcon2019pytorch} for implementing our models. PL enables effortless scaling to hundreds of GPUs, with memory and speed optimizations such as half-precision training, and sharded gradients during distributed training.  We add memory-mapped data loaders, which allow us to scale to datasets with hundreds of millions of query-passage pairs.  We use pre-trained encoders provided by the Huggingface transformers library~\citep{wolf-etal-2020-transformers}.

\paragraph{In-batch negatives}
Following \cite{dpr} we implement in batch negatives by using the differentiable all gather primitive provided by PL. Unlike the original implementation in \cite{dpr} this lets us gather negatives across all nodes leading to higher training efficiency.

\paragraph{Validation Metrics}
Evaluating neural retrieval models require embedding tens of millions of passages for indexing.  This is a one-time, manageable cost for deployment systems, however for research iteration and model selection purposes, it is prohibitively expensive.  One option is to use proxy-metrics such as validation cross-entropy loss, or in-batch accuracy to do model selection.  Unfortunately such metrics often do not correlate well with end-to-end retrieval accuracy.  As a middle-ground, we implement distributed in-memory validation using the all gather primitive. This allows us to use a fairly large proxy corpus of up to 300k passages, including up to 50 hard negative examples for each test query.  We find that using mean reciprocal rank on this corpus as a model selection metric correlates well with full evaluation metrics.

\paragraph{Training details}
Pre-training on PAQ and Reddit are run for up to 10 epochs on 64 Nvidia V100 32GB GPUs, with ADAM optimizer and triangular learning rate schedule.  Learning rate and batch size vary for each model, and are presented in the appendix. We fine-tune for up to 40 epochs on the end task on 8 GPUs.  For BERT and DeBERTa models, we use [CLS] token directly as representation, whereas for RoBERTa we add a linear projection of the same size and an additional layer normalization. BERT models use seperate encoders for query and passage, where RoBERTa and DeBERTa models use shared encoders.  

\paragraph{Data preparation}
For PAQ pre-training, we mined negative examples using a publicly available DPR checkpoint.  For Reddit, the engineering effort to setup an index of 200M documents was too large, therefore we pre-train without negatives.  For MSMARCO and KILT, we use standard pre-processing and splits, and for NaturalQuestions we follow~\citep{dpr}.  For the dialogue tasks, we use the dataset-provided negative examples when available.  We concatane all dialogue context (including persona for ConvAI2) to form the query, and truncate from the beginning if longer than 256 tokens.  

We will publicly release all code and pre-trained checkpoints for the use of the community.

\begin{table*}
\centering
\resizebox{\textwidth}{!}{   
\fontsize{8.4}{10.1}\selectfont \setlength{\tabcolsep}{0.5em}
\begin{tabular}{cccccc}
\toprule         
\multirow{2}*{\textbf{Methods}}         & 
\multirow{2}*{\textbf{Base model}} & 
\multicolumn{1}{c}{\textbf{MSMARCO}} & \multicolumn{3}{c}{\textbf{Natural Questions}} \\
                    &  & MRR@10    & R@5         & R@20         & R@100         \\
\hline
BM25 (anserini)~\citep{yang2017anserini} & -    & 18.7 & - &  59.1   & 73.7 \\
\hline
DPR (single)~\citep{dpr} & BERT$_\text{base}$       & -    & 65.8 & 78.4 & 85.4 \\
GAR~\citep{mao2020generation} & -        & -      & - & 74.4    & 85.3 \\
ANCE (single)~\citep{xiong2020approximate} & RoBERTa$_\text{base}$ & 33.0 & - & 81.9 & 87.5 \\
RocketQA~\citep{qu2021rocketqa} & ERNIE$_\text{base}$                                               & 37.0     &  74.0      & 82.7    & 88.5      \\
\hline 
DPR(ours)  & BERT$_\text{base}$ & 29.0 & 65.5 &78.3 &85.6 \\
DPR(ours)  & BERT$_\text{large}$ & 28.8 &69.14&80.19&86.73 \\
DPR(ours)  & RoBERTa$_\text{base}$ & 29.5 & 67.00 &79.03 &85.42 \\
DPR(ours)  & RoBERTa$_\text{large}$ & 30.2 &69.67 &81.27 &87.01  \\
DPR(ours)  & DeBERTa$_\text{xlarge-v2}$ & - & 72.66 & 82.38 & 87.56 \\
\hline
DPR-PAQ  & BERT$_\text{base}$ & 31.4 & 74.5 & 83.7 & 88.6 \\
DPR-PAQ  & BERT$_\text{large}$ & 31.1 & 75.3 & 84.4 & 88.9 \\
DPR-PAQ  & RoBERTa$_\text{base}$ & 32.3 & 74.15 & 84.01 & 89.2 \\
DPR-PAQ  & RoBERTa$_\text{large}$ & 34.0 & 76.93 &84.68  & 89.22  \\
DPR-PAQ  & DeBERTa$_\text{xlarge-v2}$ & - & 73.38 & 83 & 88.61 \\
\bottomrule
\end{tabular}
}
\caption{Passage retrieval results for MSMARCO development set and NaturalQuestions test set. }
\label{tab:ir}
\vspace{-4mm}
\end{table*}
\begin{table*}[t]
\centering

\resizebox{\textwidth}{!}{   
\fontsize{8.4}{10.1}\selectfont \setlength{\tabcolsep}{0.5em}
\begin{tabular}{llrrrrrrrrr}
 \toprule
 \textbf{Methods}  & \textbf{Base model} & \textbf{FEV} & \textbf{T-REx} & \textbf{zsRE}  & \textbf{NQ} & \textbf{HoPo} & \textbf{TQA} & \textbf{WoW} & \textbf{Avg.}  \\
\midrule
BM25 & - & 40.1 & 51.6 & 53.0 & 14.2 & 38.4 & 16.2 & 18.4 & 33.1 \\
Multi-task DPR & BERT$_\text{base}$ & 52.1 & 61.4 & 54.1 & 40.1 & 41.0 & 34.2 & 24.6 & 43.9 \\

\midrule
DPR-PAQ & BERT$_\text{base}$ & 61.4 & 68.4 & 73.28 & 44.1 & 44.6 & 38.9 & 26.5 & 50.6 \\
DPR-PAQ & BERT$_\text{large}$ & 62.8 & 66.58 & 66.9 & 42.6 & 42.1 & 37.9 & 23.4 & 48.9 \\
\bottomrule
\end{tabular}
}
\caption{Paragraph-level $R$-Precision on the KILT benchmark.}
\label{tab:kilt}
\end{table*}
\section{Main results}
In this section, we summarize our main results, before we dive into some analysis in the next section.  

\subsection{Passage retrieval results}\label{sec:ir_results}
Our main results for passage retrieval are presented in two tables, Table~\ref{tab:ir} for MSMARCO and NQ, and Table~\ref{tab:kilt} for KILT.  PAQ-based pre-training results in strong gains on almost all passage retrieval tasks.  For NaturalQuestions, pre-training improves +3.2 points over our non-pretrained baseline on top-20 accuracy, without using iterative training (Figure~\ref{fig:ir-data-ablation}).  Our setup with iterative training is most similar to~\citep{xiong2020approximate}, on which pre-training improves by \textit{additional} 2.1 points (81.9 vs. 84.0).  We advance the best published results~\citep{qu2021rocketqa} by +1.7 points on both top-20 and top-100 accuracy.  We note that the main contribuition of \citep{qu2021rocketqa} is using a large cross-encoder model to pre-filter training data - an approach which is orthogonal to pre-training and could provide additional gains.  On MSMARCO, we see similar gains, improving +3.8 points over our best non-pretrained baseline.  

On KILT, we advance passage retrieval SoTA on all tasks by 6.7 points of R-precision on average.  This result shows that PAQ-based pre-training generalizes well across a wide variety of tasks.  

\begin{table*}
\centering
\resizebox{\textwidth}{!}{   
\fontsize{8.4}{10.1}\selectfont \setlength{\tabcolsep}{0.5em}
\begin{tabular}{cccccccc}
\toprule         
\multirow{2}*{\textbf{Methods}}         & 
\multirow{2}*{\textbf{Base model}} & 
\multicolumn{1}{c}{\textbf{ConvAI2}} & \multicolumn{2}{c}{\textbf{DSTC7}} & \multicolumn{2}{c}{\textbf{Ubuntu v2}} \\
                    &  & R@1          & R@1        & MRR   &     R@1  &  MRR \\
\hline
~\citep{wolf2019huggingface} & BERT$_\text{base}$ & 82.1 & - & - & - & - \\
~\citep{chen2019sequential}  & BERT$_\text{base}$ & - & 64.5 & 73.5 & - & - \\
~\citep{dong2018enhance}  & BERT$_\text{base}$ & - & - & - & 75.9 & 84.8 \\
~\citep{humeau2019poly} & BERT$_\text{base}$ & 83.3 & 66.8 & 74.6 & 80.6 & 88.0 \\
~\citep{humeau2019poly} (Reddit) & BERT$_\text{base}$ & 86.9 & 70.9 & 78.1 & 83.6 & 90.1 \\
\hline
DPR (ours) & BERT$_\text{base}$  & 82.4 & 53.1 & 62.6 & 80.6 & 87.9 \\
DPR (ours) & RoBERTa$_\text{base}$  & 84.6 & 58.4 & 68.2 & 84.2 & 90.4 \\
\hline
DPR-Reddit & BERT$_\text{base}$ & 88.5 & 61.5 & 70.2 & 82.0 & 88.8 \\
DPR-Reddit & BERT$_\text{large}$ & 88.2 & 62.0 & 70.9 & 81.8 & 88.7 \\
DPR-Reddit & RoBERTa$_\text{base}$ & 88.4 & 66.5 & 75.1 & 85.1 & 90.9 \\
DPR-Reddit & RoBERTa$_\text{large}$ & 90.7 & 68.2 & 76.4 & 86.3 & 91.7 \\
\bottomrule
\end{tabular}
}
\caption{Dialogue retrieval results. }
\label{tab:dialogue}
\vspace{-4mm}
\end{table*}
\subsection{Dialogue retrieval results}\label{sec:dialogue_results}
To further verify the matched-domain hypothesis, we conduct experiments in the dialogue retrieval domain, using Reddit chat threads as pre-training data.  We see clear gains on all datasets over vanilla BERT baselines, affirming the usefulness of additional pre-training for retrieval.  However, the gains are less pronounced for UbuntuV2, which has a much larger training dataset.  Nevertheless, our best model (RoBERTa$_{large}$) still outperformes the previous SoTA by a comfortable margin on two tasks.  For DSTC7, the results also support our conclusions, however we were not able to reproduce previous baselines on this dataset, and our numbers are generally lower.

\begin{table}
\centering

\begin{tabular}{lcc}
 \toprule
 \textbf{Pre-training data} & \textbf{w/o FT} & \textbf{w/ FT}  \\
\midrule
None & - & 78.4 \\
BFS & 37.0 & 75.7 \\
ICT & 25.5 & 77.0 \\
PAQ & 78.1 & 81.6 \\

\bottomrule
\end{tabular}
\caption{Comparison of different pre-training data, with and without fine-tuning (FT). Metric is top-20 accuracy on NaturalQuestions test set. Baseline is vanilla BERT-base model.}
\label{tab:pretraining}
\end{table}

\section{Pre-training retrieval models}

In this section we cover our findings regarding how to best pre-train bi-encoder models for retrieval.  We compare our pre-training approach with previous approaches, and emphasize the importance of picking the right pre-training task.  We discuss the effects of data and model size for pre-training retrieval models.

\subsection{Picking the pre-training task}
As pointed out earlier, previous attempts at pre-training dense retrieval models have largely been ineffective.  In Table~\ref{tab:pretraining}, we confirm this conclusion.  We see that BFS and ICT do result in non-trivial zero-shot retrieval performance on the NQ dataset. However, after fine-tuning these gains disappear, and they do not outperform a vanilla BERT model.  
The performance of PAQ-pretrained retrieval is exceptionally strong even before fine-tuning.  This is expected to an extent, since PAQ has been trained on NQ, and many NQ training questions might already appear verbatim in the PAQ generated questions.  Nevertheless, pre-training with PAQ results in robust gains, which persist after fine-tuning.  Note that both BFS and ICT were pre-trained on more data than PAQ (200 million pairs vs. 65 million).  We conclude that PAQ pairs are higher quality, and better matched to the end task than previously proposed artificial pre-training tasks, resulting in better performance.

For the dialogue experiments, we compare against \cite{humeau2019poly}, who also pre-trains on the same Reddit corpus, but using a cross-encoder with masked-language-modeling and next-sentence-prediction objectives a la BERT~\citep{bert}.  This allows us to compare bi-encoder pre-training, with cross-encoder pre-training on the same dataset.  Looking at Table~\ref{tab:dialogue}, we see that bi-encoder pre-training (DPR-Reddit, BERT$_{base}$) performs significantly better than cross-encoder pre-training on the ConvAI2 dataset.  However, the same conclusion does not hold for the larger and more domain-mismatched Ubuntu corpus. (Our RoBERTa-large bi-encoder does improve over~\citep{humeau2019poly}, however we don't have a corresponding cross-encoder pre-trained baseline for this model.) We conclude that transfer is somewhat fragile for dense retrieval pre-training, and is sensitive to domain and task mismatch.

\subsection{Effect of data size}
In Figure~\ref{fig:ir-data-ablation} we investigate the effect of pre-training data size on retrieval performance.  We randomly downsample the PAQ pre-training dataset, and plot top-20 accuracy on NQ after fine-tuning on the full NQ training set.  We see that as little as 1 million pre-training examples can improve performance, with larger pre-training data resulting in more gains as expected.  This suggests expanding PAQ with even more questions could potentially be beneficial (though this could be contingent on the quality of additional generated questions).
\begin{figure}
    \centering
    \includegraphics[scale=0.5]{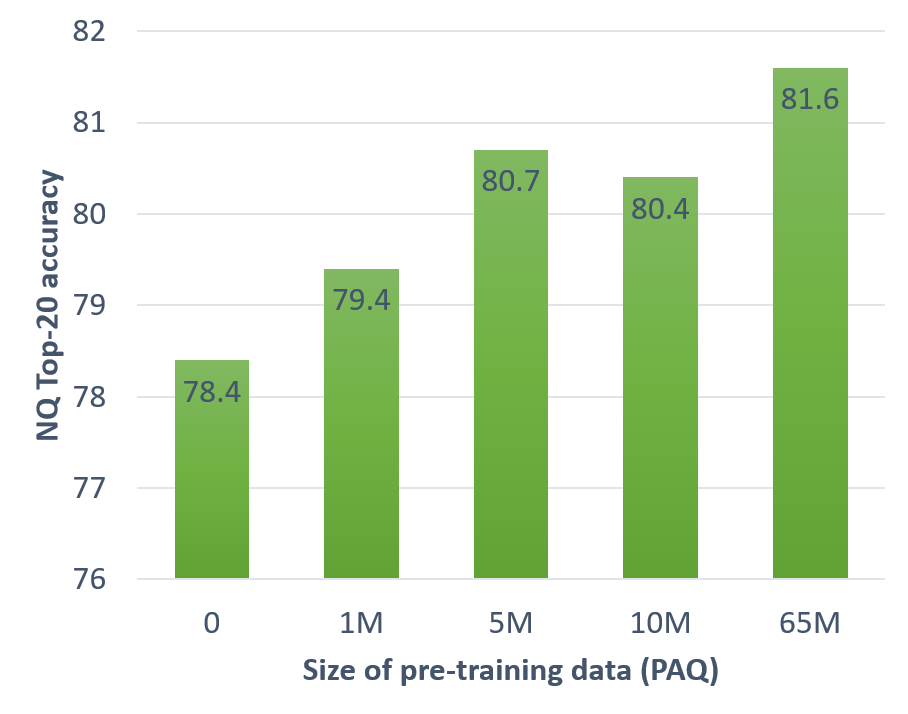}
    \caption{Effect of different sizes of PAQ data for pre-training.  Results show top-20 accuracy on NQ after fine-tuning.  No iterative pre-training is used.}
    \label{fig:ir-data-ablation}
\end{figure}

\ignore{
\begin{figure}[ht]
    \centering
        \begin{tikzpicture}[scale=0.9\tikzscale]
        \begin{axis}[
            xmode=log,
            log ticks with fixed point,
            xlabel={Training data size (millions)},
            ylabel={IR top-20 Accuracy (\%)},
            xmin=0, xmax=70,
            ymin=75, ymax=85,
            xtick={0, 1, 5, 10, 65},
            ytick={10,20,30,40,50,60,70,80,90},
            legend pos=south east,
            ymajorgrids=true,
            grid style=dashed,
        ]

        \addplot[
            color=blue,
            mark=square,
            mark size=1pt
            ]
            coordinates {
            (0.01, 78.4)(1,79.4)(5,80.7)(10,80.4)(65,81.6)
            };
            \addlegendentry{fine-tuned}

        \end{axis}
        \end{tikzpicture}

    \caption{Top-20 accuracy on NQ.}
    \label{fig:ir-data-ablation}
\end{figure}
}

It's interesting to note that additional MLM training is not generally helpful for retrieval on open-domain QA.  RoBERTa~\citep{RoBERTa} was trained on an order of magnitude more data for much longer compared to BERT, yet fine-tuning on RoBERTa results in little, if any improvement over BERT, in the absence of retrieval-specific pre-training (Table~\ref{tab:ir}).  For dialogue retrieval, better MLM training does help, as was shown previously~\citep{humeau2019poly}.

\subsection{Effect of model size}
We experimented with pre-trained models of varying sizes, including BERT(base/large)~\citep{bert}, RoBERTa(base/large)~\citep{RoBERTa}, and DeBERTa(xlarge-v2)~\citep{he2020deberta}.  In terms of how better and larger pre-trained models interact with retrieval-specific pre-training, we get mixed results.  For instance, for passage retrieval (Table~\ref{tab:ir}), DeBERTa-xlarge-v2 model does outperform the BERT-base baseline significantly in the fully supervised setting, yet this gain disappears after additional pre-training with PAQ.  The opposite is true when comparing RoBERTa vs. BERT, as we see RoBERTa performing better after intermediate pretraining, both for passage retrieval and for dialogue tasks (Table~\ref{tab:dialogue}).  In contrast to the clear-cut conclusions for other NLP tasks, it is hard to conclude whether larger and better language models actually make better retrieval models.

\subsection{Effects of PAQ on Retrieval}
In section \ref{sec:ir_results}, we established that pretraining on PAQ is beneficial for passage retrieval for QA. 
However, it is worth considering where the source of this improvement lies.
\citet{lewis2021paq} note that QA-pairs in PAQ have substantial overlap with the test sets of NQ --- indeed, this is intentional, given their aim of preempting a large number of probable questions for use as a cache for question answering models.
In fact, $\sim$9\% of the NQ test questions appear verbatim in PAQ. 

It is worth investigating then, whether the gains we observe are due to simply memorising the relevant passages for PAQ questions which overlap with test questions, or, whether they are due to learning more robust, generalizable model behaviour.

\begin{table}
\centering

\begin{tabular}{lcc}
 \toprule
  \backslashbox[25mm]{\footnotesize{DPR}}{\footnotesize{DPR-PAQ}} & R@20 \cmark & R@20 \xmark
  \\
\midrule
R@20 \cmark & 2.5 & 3.2\\
R@20 \xmark & 3.3& 3.1\\
\bottomrule
\end{tabular}
\caption{Mean Levenshtein distance to most similar question in PAQ, for DPR-PAQ and a DPR baseline for NQ test questions, stratified by whether the model achieves Recall@20}
\label{tab:paq_analysis}
\end{table}
To investigate, we compare the predictions of the DPR-PAQ retriever with an otherwise equal baseline DPR model, without PAQ-pretraining.
On the subset of the NQ test set that overlaps verbatim with PAQ questions, we find that DPR-PAQ achieves 95.5\% R@20, whereas the baseline achieves 94.8\%
These are both remarkably high scores, indicating that these verbatim questions are very easy for models to solve, regardless of pretraining. 
Due to the very similar performance on this subset, the difference in overall performance cannot be attributed to simply memorising verbatim-overlapping questions.

To investigate further, we perform an analysis of the NQ test questions for which the DPR-PAQ model at least one correct passage in the top 20, but our baseline DPR model does not, and vice versa. 
To determine whether these test questions are similar to questions seen in pretraining, we retrieve the 100 most similar questions in PAQ for each test question using the RePAQ question retriever from \citet{lewis2021paq}, and calculate the minimum Levenshtein edit distance.  

Table \ref{tab:paq_analysis} shows these results.
What we find is that test questions that both the DPR baseline and DPR-PAQ  retrieve well for are the most similar to PAQ questions, with a mean minimum edit distance of 2.5 words.
%
%
We also note that questions that one model retrieves well for but the other does not tend to have high edit distances to PAQ. (3.2 and 3.2 words for DPR and DPR-PAQ respectively).
If DPR-PAQ's improvement was due to simple PAQ memorization, we would not expect to observe these high edit distances on questions for which it outperforms the baseline.
This suggests the improvement cannot be explained by simply memorising PAQ, and that PAQ-pretraining improves retrieval performance in a more general way.  This is corroborated by the strong results obtained with PAQ pre-training on the KILT and MSMARCO benchmarks as discussed above.

\section{Related Work}

\subsection{Dense retrieval}
\citet{orqa} was first to show that dense pre-trained representations can outperform BM25 for end-to-end retrieval in the context of open-domain QA.  This work also proposed the ICT pre-training task for retrieval, and demonstrated its usefulness.  \citet{REALM} improved on this work, by end-to-end pre-training of retriever and reader using a language modeling loss.  It was subsequently shown \citep{dpr} that these sophisticated end-to-end pre-training methods are not necessary, and a fully-supervised fine-tuning of the retriever can produce superior results.  The performance of fully-supervised models were improved even further in \citep{xiong2020approximate} and \citep{qu2021rocketqa} by iteratively updating negative candidates, using cross-encoder models for increasing the quality of negative candidates, and hyperparameter optimizations.
\paragraph{Pretraining for retrieval}
\citet{chang2020pre} investigate several artificial tasks for training dense retrieval models, including ICT and BFS, showing improvements over no pre-training.  However their setting is not fully open, and they report on a smaller set of 1 million passages.  These results have also been superseded by better supervised fine-tuning.  

Concurrent work \citep{sachan2021end} combined ICT pre-training with masked-salient-span pre-training, as well as an end-to-end fine-tuning using a T5-large model, obtaining results comparable or slightly better than what's presented here.  The major improvements in this work are attributed to end-to-end training, which amounts to a type of distillation from the powerful T5 model into the retrieval model.  It's interesting to compare this to more direct distillation methods \citep{izacard2020distilling, yang2020retriever}, which also reported similar gains.  Our method also relies on a reader model indirectly, through the global filtering stage of generated questions in PAQ.  However, this is different and more general than mere distillation on a supervised dataset, as it also involves data augmentation at large scale, and generalizes well to other datasets, as shown in section~\ref{sec:ir_results}. 

\paragraph{Question generation}
\citet{lewis2021paq}, which we heavily rely on, used generated questions as a cache to build a fast lookup-based QA system.  Using the same question bank as pre-training, we have shown that we can get additional value and generalisation from this resource.  \citet{ma2021zero} and \citet{jia2021question} also investigate training on generated QA pairs, but the former only considers application to domain transfer and the latter to other NLP tasks.

\section{Conclusion}
We have investigated domain-matched pre-training tasks for bi-encoder dense retrieval models.  We found that the proposed approach is more effective than previously proposed artificial pre-training tasks.  We demonstrated the generality of our conclusions, by evaluating on a large and varied set of passage retrieval and dialogue retrieval benchmarks. 

Our work should be considered as a new statement in the ongoing dialogue of how to best train dense retrieval models.  We believe we have addressed some important open questions, such as whether and when pre-training can be useful.  However we have also raised new questions, in addition to the many which remain open.  For instance, many different ways of leveraging reader models for better retrieval have been recently proposed, including end-to-end training, distillation, data filtering and data augmentation.  What is the relationship between these approaches?  Are they complementary?  Which ones are more efficient, and more performant?  We believe these questions deserve a more thorough investigation.

We have focused on mostly on dense retrieval when full supervision is available, and showed that for $k=100$ retrieval candidates, the performance is already approaching a ceiling.  There is more room for improvement for smaller $k$.  In this regime however, re-ranking models also become feasible and separable architecture is not a strict requirement.  Therefore, further improvements to retrieval will likely need to be discussed with more emphasis on the computation-accuracy trade-off.  Few-shot and zero-shot retrieval will also be of increasing importance, and there are already works investigating this direction~\citep{maillard2021multi,thakur2021beir}.

\bibliography{acl2021}
\bibliographystyle{acl_natbib}

\appendix

\section{Appendices}
\label{sec:appendix}
\subsection{Pre-training hyperparameters}

\begin{table}[h!]
\centering
\begin{tabular}{lcc}
 \toprule
 \textbf{Encoder} & \textbf{lr} & \textbf{bs}  \\
\midrule
BERT$_{base}$ & 2.5e-5 & 32 \\
BERT$_{large}$ & 1e-5 & 12 \\
RoBERTa$_{base}$ & 2e-5 & 40 \\
RoBERTa$_{large}$ & 1e-5 & 12 \\
DeBERTa$_{xlarge}$ & 1e-5 & 12 \\

\bottomrule
\end{tabular}
\caption{Learning rate and batch size for pre-training.}
\label{tab:hyperparams}
\end{table}

\end{document}